\documentclass[runningheads]{llncs}
\usepackage{graphicx}
\usepackage{tikz}
\usepackage{comment}
\usepackage{amsmath,amssymb} % define this before the line numbering.
\usepackage{color}

\usepackage{booktabs}

\usepackage{algorithm}
\usepackage{algorithmic}
\usepackage{epsfig}
\usepackage{framed,multirow}
\usepackage{latexsym}
\usepackage{cite}
\usepackage{xcolor}
\usepackage{multirow}
\usepackage[normalem]{ulem}
\usepackage{booktabs}
\usepackage{arydshln}

\usepackage{url}

\newcommand{\mb}[1]{\mathbf{#1}}

\newcommand{\tb}[1]{\textbf{#1}}
\newcommand{\tu}[1]{\underline{#1}}
\usepackage[accsupp]{axessibility}  % Improves PDF readability for those with disabilities.

\begin{document}
% \linenumbers
\pagestyle{headings}
\mainmatter
\def\ECCVSubNumber{5400}  % Insert your submission number here

\title{Out of Sight, Out of Mind:\\A Source-View-Wise Feature Aggregation for Multi-View Image-Based Rendering} % Replace with your title

\begin{comment}
\titlerunning{ECCV-22 submission ID \ECCVSubNumber} 
\authorrunning{ECCV-22 submission ID \ECCVSubNumber} 
\author{Anonymous ECCV submission}
\institute{Paper ID \ECCVSubNumber}
\end{comment}
%******************

%\begin{comment}
\titlerunning{ }
\author{Geonho Cha\inst{1} \and
Chaehun Shin\inst{2} \and
Sungroh Yoon\inst{2, 3} \and Dongyoon Wee\inst{1}}
\authorrunning{G. Cha et al.}
\institute{Clova AI, NAVER Corp., Korea \email{\{geonho.cha,dongyoon.wee\}@navercorp.com}\and
Department of Electrical and Computer Engineering, Seoul National University, Korea
\email{\{chaehuny,sryoon\}@snu.ac.kr}\and
Interdisciplinary Program in Artificial Intelligence, Seoul National University, Korea}
%\end{comment}
%******************
\maketitle

\begin{abstract}
To estimate the volume density and color of a 3D point in the multi-view image-based rendering, a common approach is to inspect the consensus existence among the given source image features, which is one of the informative cues for the estimation procedure.
To this end, most of the previous methods utilize equally-weighted aggregation features.
However, this could make it hard to check the consensus existence when some outliers, which frequently occur by occlusions, are included in the source image feature set.
In this paper, we propose a novel source-view-wise feature aggregation method, which facilitates us to find out the consensus in a robust way by leveraging local structures in the feature set.
We first calculate the source-view-wise distance distribution for each source feature for the proposed aggregation.
After that, the distance distribution is converted to several similarity distributions with the proposed learnable similarity mapping functions.
Finally, for each element in the feature set, the aggregation features are extracted by calculating the weighted means and variances, where the weights are derived from the similarity distributions.
In experiments, we validate the proposed method on various benchmark datasets, including synthetic and real image scenes.
The experimental results demonstrate that incorporating the proposed features improves the performance by a large margin, resulting in the state-of-the-art performance.
\end{abstract}

\section{Introduction}
\label{sec:intro}
Novel view synthesis is one of the core problems in computer vision, which has various applications like augmented reality and human-computer interaction.
Various methods based on explicit scene representations \cite{penner:soft3d:acm2017, zhou:stereomagni:siggraph2018, srinivasan:mpiextra:cvpr2019, mildenhall:llff:acm2019, flynn:deepview:cvpr2019, tucker:sigleviewmpi:cvpr2020, sitzmann:deepvoxels:cvpr2019, lombardi:neuralvolumes:arxiv2019} or implicit scene representations \cite{liu:nsvf:neurips2020, niemeyer:dvr:cvpr2020, saito:pifu:cvpr2019, sitzmann:srn:neurips2019, yariv:multiview:neurips2020} have been proposed to handle this problem.
Recently, neural radiance fields (NeRF) \cite{mildenhall:nerf:eccv2020}, which is based on implicit scene representation, has shown promising results in the novel view synthesis problem.
In order to synthesize the image from a novel view, NeRF optimizes multi-layer perceptron (MLP) with multi-view images of the scene.
Though per-scene optimization helps to generate fine-detailed images, it hinders practical usage due to its expensive optimization cost.

% generalizable nerf
To mitigate the expensive optimization cost, multi-view image-based rendering methods \cite{yu:pixelnerf:cvpr2021, wang:ibrnet:cvpr2021, chen:mvsnerf:arxiv2021, trevithick2021grf} have been introduced to exploit 2D features from source images of the scene during the rendering.
They utilize aggregated 2D features to predict the densities and colors of corresponding sampled 3D points.
Thus, how to aggregate this feature set is crucial for precise density and color prediction for these methods.
To aggregate 2D features, previous multi-view image-based rendering methods typically utilize element-wise global mean and variance, based on the observation that 3D points on surfaces are more likely to have consensus in the 2D feature set.
Accordingly, a network learns to reason density by comparing features with global mean based on variance.

%From the source images, they extract features to comprise feature set, which is used to estimate the volume density and color of each sampled point on the ray.
%Inspired by the classical multi-view 
%The estimation process is designed inspired by the classical multi-view stereo method \cite{scharstein2002taxonomy, szeliski2007image}, i.e., the 3D point is likely to be on a surface if there is a consensus in the feature set.

\begin{figure}[t]
    \centering
    \includegraphics[height=5.1cm]{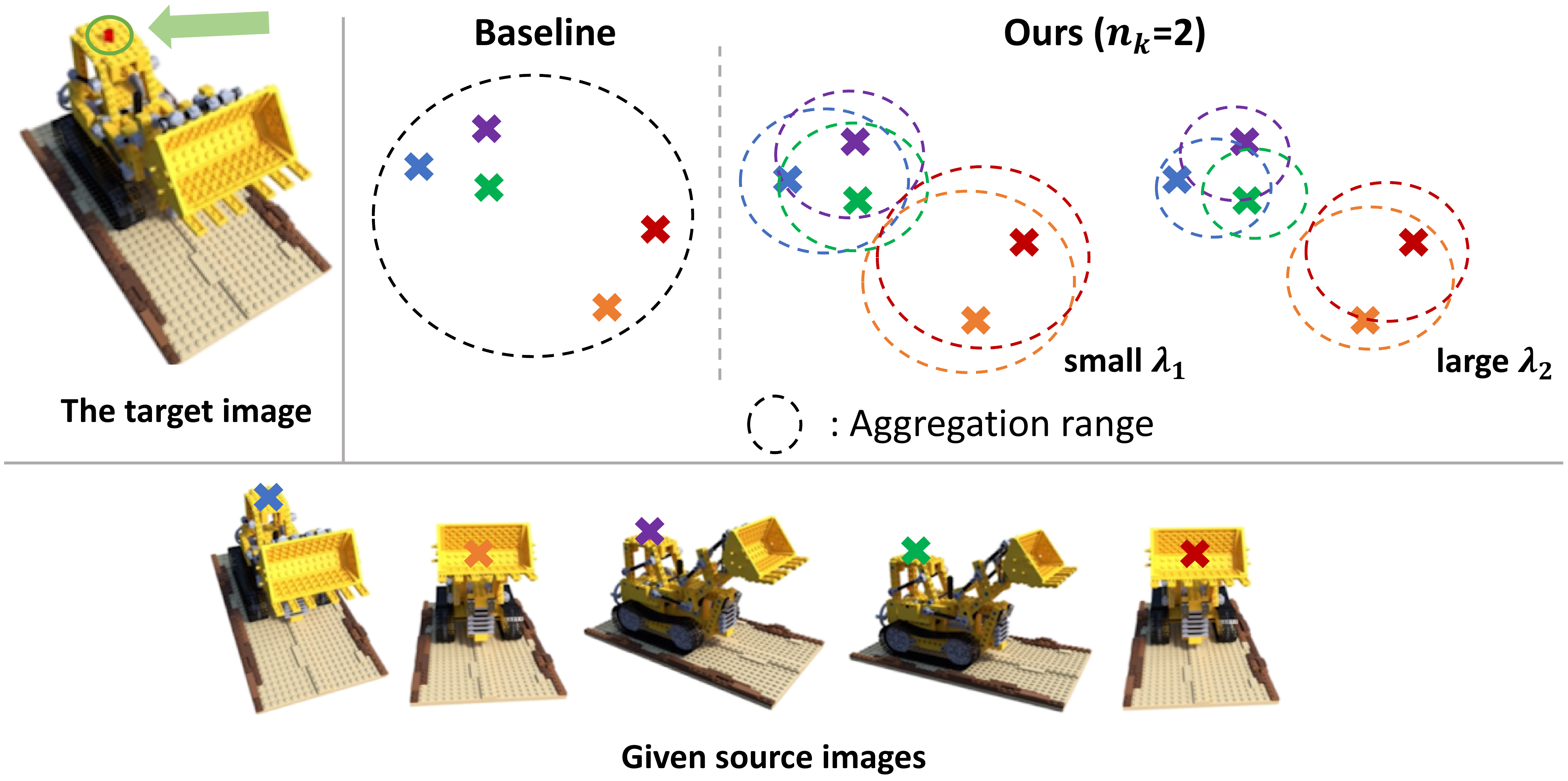}
    \vspace{-2mm}
    \caption{An illustration of the proposed method's intuition. Let us consider a rendering process of the target image pixel indicated by the green arrow, given the corresponding source image features indicated by `x' marks. The baseline methods \cite{wang:ibrnet:cvpr2021, yu:pixelnerf:cvpr2021, chen:mvsnerf:arxiv2021} utilize a single aggregation range including outlier features so that outlier features influence to get mean and variance of features. Meanwhile, the proposed method uses multiple source-view-wise aggregation ranges, which can suppress the influence of the outlier features. This enables us to figure out the consensus existence in a robust way. Here, the learnable parameter $\lambda$ regulates the aggregation range.}
    \label{fig:teaser}
\end{figure}

%To figure out the consensus existence, most of the previous methods \cite{wang:ibrnet:cvpr2021, yu:pixelnerf:cvpr2021, chen:mvsnerf:arxiv2021} employ globally aggregated features like element-wise mean and variance of the features in the feature set.
%In other words, there is a consensus in the feature set and an element in the feature set is likely to be in the consensus if it has similar feature values to the global mean and the global variance has low values.
A thing to note here is that they give the same weight to each 2D feature when calculating the aggregation features.
However, the equally-weighted aggregation features could not be informative in situations which some outlier features, which could frequently occur with occlusions, are in the feature set.
For example, in Figure \ref{fig:teaser}, even though the 3D point is on a surface, features extracted from the occluded region of the bucket would be different from the features extracted from the region of the ceiling.
In this case, the equally-weighted aggregation features are not suitable for finding a consensus in the feature set.

%finding out the consensus existence based on the globally aggregated features could degrade the performance (the synthesized image could have blurs or artifacts) in situations which some outlier features, which could frequently occur with occlusions, are in the feature set.
%For example, in Figure \ref{fig:teaser}, even though the 3D point is on a surface, features extracted from the occluded region of the bucket would be different from the features extracted from the region of the ceiling.
%In this case, globally aggregated features are not suitable for finding a consensus in the feature set as they aggregate the features with the same weights.

% local feature proposed
In this paper, to resolve this issue, we propose a novel source-view-wise feature aggregation method, which facilitates us to find out the consensus existence in a robust way utilizing the local structure of the feature set.
Unlike the previous methods that use common equally-weighted aggregation features, our approach incorporates multiple aggregation features which are aggregated based on source-view-wise weights.
For the proposed aggregation, we first calculate the source-view-wise distance distribution for each element in the feature set.
Thereafter, the distance distribution is converted to similarity distributions through the proposed learnable similarity measure functions.
Here, the learnable similarity measure functions are trained in an end-to-end manner with other network parameters to be tailored to the 2D features.
Finally, we extract the aggregation features by calculating the weighted element-wise means and variances, where the weights are determined based on the similarity distributions.
These features are utilized along with the source image features to infer the volume densities and colors of 3D points in the volume rendering process.

% experimental
In experiments, we validate the proposed method on several benchmark datasets consisting of synthetic and real scenes.
The experimental results show that incorporating the proposed feature set improves the performance by a large margin, which demonstrates the effectiveness of the proposed source-view-wise feature aggregation method.

% contribution
The contributions of our method can be summarized as:
\begin{itemize}
    \item We propose a novel source-view-wise feature aggregation method which facilitates us to utilize the consensus existence in a robust way by leveraging the local structures in the feature set.
    \item We propose novel learnable similarity measure functions for the proposed feature extraction.
    \item The proposed scheme shows the state-of-the-art performance on the diverse benchmark datasets.
\end{itemize}

\section{Related work}
\label{sec:related}
The novel view synthesis problem has been actively dealt with as it has various applications.
The methods handling the problem can be categorized into two types based on how they model 3D spaces.
The first type models the 3D space with explicit scene representations \cite{penner:soft3d:acm2017, zhou:stereomagni:siggraph2018, srinivasan:mpiextra:cvpr2019, mildenhall:llff:acm2019, flynn:deepview:cvpr2019, tucker:sigleviewmpi:cvpr2020, sitzmann:deepvoxels:cvpr2019, lombardi:neuralvolumes:arxiv2019, kalantari:learning:acmtog2016, kar:learning:neurips2017, tulsiani:multi:cvpr2017, flynn:deepstereo:cvpr2016, li:crowdsampling:eccv2020}, and the second one models the 3D space based on implicit scene representations \cite{liu:nsvf:neurips2020, niemeyer:dvr:cvpr2020, saito:pifu:cvpr2019, sitzmann:srn:neurips2019, yariv:multiview:neurips2020}.

The explicit-representation-based methods utilize voxel grids \cite{lombardi:neuralvolumes:arxiv2019, kalantari:learning:acmtog2016, kar:learning:neurips2017, penner:soft3d:acm2017, tulsiani:multi:cvpr2017, sitzmann:deepvoxels:cvpr2019} or multi-plane images \cite{flynn:deepview:cvpr2019, flynn:deepstereo:cvpr2016, li:crowdsampling:eccv2020, mildenhall:llff:acm2019, srinivasan:mpiextra:cvpr2019, zhou:stereomagni:siggraph2018, tucker:sigleviewmpi:cvpr2020} to perform the novel view image synthesis process.
However, these methods have some drawbacks: (i) it is hard to infer high-resolution images as the required memory size increases drastically with the increased resolutions because of the explicitly inferred 3D spaces, and (ii) some artifacts could occur comes from the discretized nature.

To handle these issues, implicit-representation-based 3D object modeling methods have been proposed \cite{park:deepsdf:cvpr2019, mescheder:occupancynetwork:cvpr2019, genova:ldif:cvpr2020, jiang:localimplicitgrid:cvpr2020}.
They modeled the 3D objects based on signed distance functions \cite{park:deepsdf:cvpr2019, jiang:localimplicitgrid:cvpr2020} or occupancy functions \cite{mescheder:occupancynetwork:cvpr2019, genova:ldif:cvpr2020}.
Incorporating these methods, implicit-representation-based novel view synthesis methods have been proposed \cite{liu:nsvf:neurips2020, niemeyer:dvr:cvpr2020, saito:pifu:cvpr2019, sitzmann:srn:neurips2019, yariv:multiview:neurips2020, liu:dist:cvpr2020}.
Recently, neural radiance fields \cite{mildenhall:nerf:eccv2020} and its following work \cite{tretschk:nonrigidnerf:cvpr2021, li:neuralff:cvpr2021, martin:nerfinwild:cvpr2021, lindell:autoint:cvpr2021, barron:mipnerf:iccv2021, yu:plenoctrees:iccv2021, rebain:derf:cvpr2021, jain:diernerf:iccv2021, liu:editnerf:iccv2021, lin:barf:iccv2021} have been proposed.
The NeRF utilizes implicit neural radiance fields, and shows solid performance on the novel view synthesis task.
They used the classical differentiable volume rendering \cite{kajiya:volumnrender:1984} to synthesize the target image, of which the volume densities and colors are estimated from a multi-layer perceptron (MLP) network.
To alleviate the issue that the MLP tends to focus on learning low-frequency information, they utilized a positional encoding technique \cite{rahaman2019spectral}.
Although its satisfactory performance, the need for per-scene optimization was a burden for their practical uses.

To relieve this condition, multi-view image-based rendering methods \cite{yu:pixelnerf:cvpr2021, wang:ibrnet:cvpr2021, chen:mvsnerf:arxiv2021, chibane:srf:cvpr2021, trevithick2021grf} have been proposed.
To render a novel view image of a scene, they utilize source image features of the same scene.
Here, figuring out the consensus existence is a key for the on-the-fly estimation of volume densities and colors.
To this end, most of the multi-view image-based rendering methods incorporated equally-weighted aggregation features.
PixelNeRF \cite{yu:pixelnerf:cvpr2021} proposed an image feature conditioned MLP to handle the generalizable novel view synthesis problem.
They aggregated the multi-view MLP outputs by an average pooling, which is fed into another MLP to estimate the volume density and color.
MVSNeRF \cite{chen:mvsnerf:arxiv2021} utilized plane-swept cost volumes for the generalized novel view synthesis.
The global variance of the source view image features was utilized to establish the cost volumes.
IBRNet \cite{wang:ibrnet:cvpr2021} proposed a MLP-based network with a ray transformer.
They utilized globally aggregated features of element-wise mean and variance.
On the other hand, GRF \cite{trevithick2021grf} utilized a geometry-aware attention module for the feature aggregation, which could be effective for the occlusion handling.
However, they did not integrate the local features and the aggregation feature.
It could be ineffective to infer surfaces which leads to inferior performance in occlusion problems.

Note here that most of the previous methods utilized equally-weighted aggregation features.
However, in this way, it could be hard to discover the surface when some outliers, which could frequently occur by occlusions, are included in the source image feature set.
In this paper, to handle this issue, we propose a novel source-view-wise feature aggregation method, which facilitates us to inspect the consensus existence in a robust way by incorporating the local structure in the feature set.

\section{Multi-view image-based rendering}
\label{sec:problem_formulation}
We handle the problem of synthesizing the target image $\mb{I}_t$ in the target camera view $\mb{p}_t$ given source images $\mb{I}_s$ along with their corresponding camera parameters $\{\mb{I}_s^i, \mb{K}_s^i, \mb{p}_s^i|i=1,\cdots,n_s\}$, where $n_s$ is the number of the source images, and $\mb{K}_s^i$ and $\mb{p}_s^i$ are the intrinsic and extrinsic camera parameters of the $i$-th source image, respectively.

For the target image rendering, we utilize the differentiable volume rendering scheme \cite{kajiya:volumnrender:1984}.
First, we shoot ray $\mb{r}(t)=\mb{o}+t\mb{d}$ for each pixel of the target image, where $\mb{o}$, $\mb{d}$, and $t$ are the target camera center, target viewing direction, and depth, respectively.
After that, we sample points $\{t_i|i=1,\cdots,n_t\}$ on the ray, where the sampling range is bounded by the given near bound $t_n$ and the far bound $t_f$.
Here, $n_t$ is the number of sampled points, and $t_u \le t_v$ if $u\le v$.
The RGB color $C$ of the pixel is rendered based on the following equation:
\begin{equation}
    C(\mb{r}) = \sum_{i=1}^{n_t}T_i(1-\textrm{exp}(-\sigma(\mb{r}(t_i))\delta_i))\mb{c}(\mb{r}(t_i),\mb{d}),
\end{equation}
where $T_i=\textrm{exp}(-\sum_{j=1}^{i-1}\sigma(\mb{r}(t_j))\delta_j)$, and $\delta_i=t_{i+1}-t_i$.
Here, $\sigma(\mb{r}(t_i))$ is the volume density of the 3D point $\mb{r}(t_i)$, and $\mb{c}(\mb{r}(t_i),\mb{d})$ is the RGB color of the 3D point $\mb{r}(t_i)$ conditioned on the viewing direction $\mb{d}$.
Note that for the volume rendering, we need to design the volume density estimator $\sigma(\cdot)$ and the color estimator $\mb{c}(\cdot)$.

In the multi-view image-based rendering approach, the features of the given source images are incorporated to infer the volume densities and colors of sampled 3D points.
Let us consider the problem of estimating the volume density and the color of a 3D point $\mb{x}$.
We first project $\mb{x}$ onto each source image as
\begin{equation}
    \mb{u}^i = \mb{\Pi}^i \mb{x},
\end{equation}
where $\mb{\Pi}^i$ is the projection matrix of the $i$-th source view image, and $\mb{u}^i$ is the projected pixel location.
We here assume homogeneous coordinates.
Based on the projected pixel locations, we compose the feature set $f=\{\mb{f}_i \in \mathbb{R}^{n_f}|i=1,\cdots,n_s\}$, which consists of the extracted source image features.
Here, $n_f$ is the feature dimension.
In this process, we get the bilinearly interpolated features $\mb{f}_i$ since the projected pixel may not have integer coordinates.
The feature set $f$ is utilized for the volume density and color estimation.

Most of the multi-view image-based rendering approaches design the estimation process after being inspired by the classical multi-view stereo scheme \cite{scharstein2002taxonomy, szeliski2007image}.
If a 3D point is on a surface, which is the case that the 3D point has a significant impact on the rendered image, there is a consensus in the feature set $f$.
The network is designed to fulfill that the features in the consensus contribute a lot when inferring the volume density and the color of the 3D point.
To achieve this, most of the previous methods utilize equally-weighted aggregation features \cite{wang:ibrnet:cvpr2021, chen:mvsnerf:arxiv2021, yu:pixelnerf:cvpr2021}.
A typical way for the aggregation is to calculate the statistical information like element-wise mean and variance of the feature set.

However, in this way, it could be hard to identify the consensus due to some outlier features which can frequently arise from occlusions.
To resolve this issue, we propose a novel source-view-wise feature aggregation method, which is introduced in the next chapter.

\section{Source-view-wise feature aggregation}
\label{sec:proposed}
\begin{figure*}[t]
    \centering
    \includegraphics[height=3.3cm]{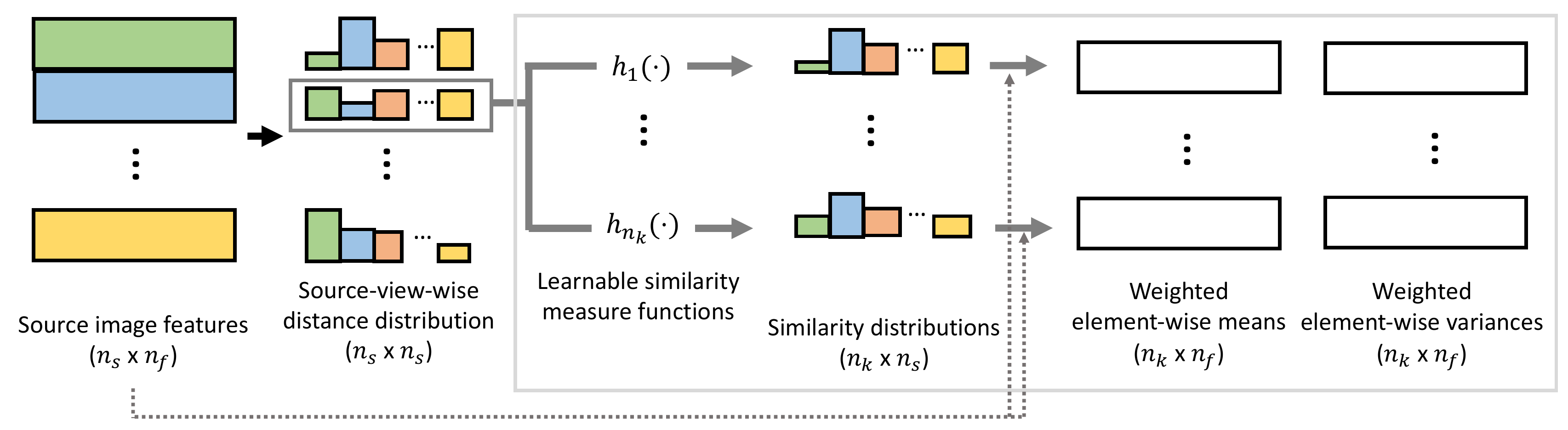}
    \caption{An overview of the proposed source-view-wise feature aggregation method. We first calculate the distance distribution for each element in the feature set for the feature aggregation. After that, the distance distributions are converted to similarity distributions with the proposed learnable similarity measure functions. Finally, the aggregated features are extracted by calculating weighted element-wise means and variances. Here, $n_s$, $n_k$, and $n_f$ are the number of source images, similarity measure functions, and the feature dimensions, respectively.}
    \label{fig:overview}
\end{figure*}

An overview of the proposed feature aggregation method is visualized in Figure \ref{fig:overview}.
Given the source image feature set $f$, the proposed method aggregates the features in a source-view-wise way to obtain aggregation features for each source feature.
We first compute the distance distribution for the $i$-th source feature as
\begin{equation}
    d_i = \{d_{i,j}|j=1,\cdots,n_s\},~d_{i,j} = \|f_i - f_j\|_2^2,
\end{equation}
where $\|\cdot\|_2$ denotes the $l_2$-norm.
After that, the distance distribution is converted to the similarity distributions based on the proposed learnable similarity measure functions.

\begin{figure}[t]
    \centering
    \includegraphics[height=4.1cm]{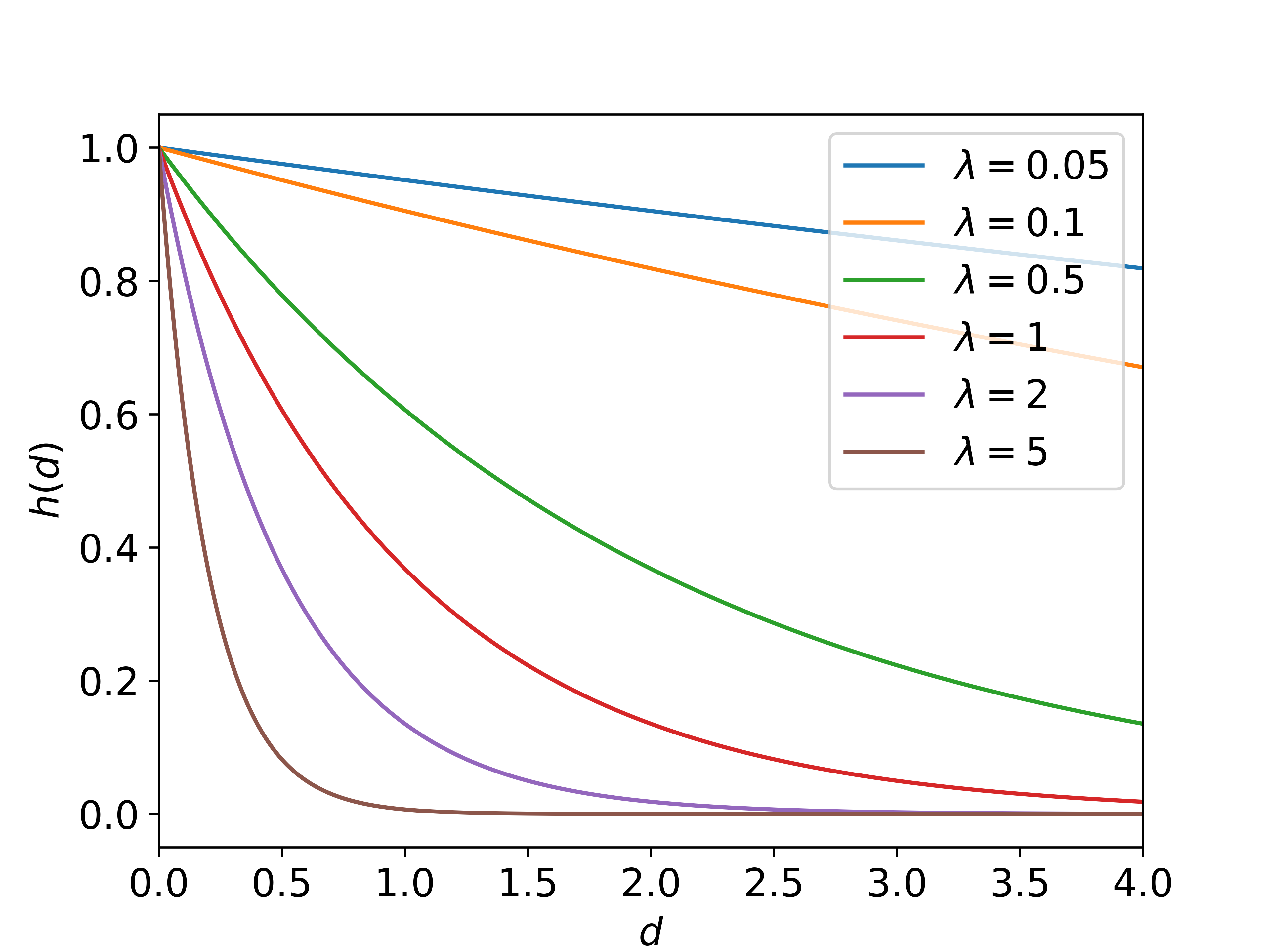}
    \vspace{-2mm}
    \caption{A graphical illustration of $h(\cdot)$ with respect to various values of $\lambda$.}
    \label{fig:graph}
\end{figure}

\subsection{Learnable similarity measure functions}
The distance distribution could be converted to a similarity distribution with a predefined function.
However, the predefined function might be a sub-optimal solution as they are not designed utilizing the extracted image features.
Moreover, it might be hard to extend the predefined function to a family of the functions, which might be needed to model the complex feature distributions.
To mitigate these issues, we propose learnable similarity measure functions.
Note that ablation studies with respect to the choice of the similarity measure functions can be found in the experiments section.

The goal of the similarity measure functions is converting the distance distribution $d_i$ to a set of similarity distributions $s_i=\{s^{k}| k=1,\cdots, n_k\}$ where $n_k$ is the number of similarity measure functions, and $s^k=\{s_{i,j}^{k}|j=1,\cdots, n_s\}$.
Here, $s_{i,j}^{k}$ is the similarity derived from $d_{i,j}$ based on the $k$-th similarity measure function.
Each similarity measure function should meet the following conditions: (i) The function should output larger similarity values in case of the smaller distances, and (ii) it should be easy to be trained, i.e., it has smooth gradient functions.
With the criteria in consideration, we propose the following functions as the similarity measure:
\begin{equation}
    h_k(d_{i,j}) = s_{i,j}^k = e^{-\lambda_k d_{i,j}},
\end{equation}
where $\lambda_k$ is a learnable scalar parameter.
The learnable parameter $\lambda_k$ is trained with the other network weights in an end-to-end manner.
Here, $\lambda_k$ regulates the aggregation range, i.e., how much the farther features are ignored.
Some illustrations of the similarity measure function with respect to various values of $\lambda$ are visualized in Figure \ref{fig:graph}.
A thing to note here is that $\lambda_k$ should have a positive value to meet the first condition of the similarity measure function.
To achieve this, we parameterize $\lambda_k$ as $e^{\alpha_k}$ where $\alpha_k$ is a learnable scalar parameter.

With the proposed method, unlike a predefined function, the similarity measure function can be designed more tailored to the extracted image features.
Furthermore, we can utilize several similarity measure functions by increasing $n_k$.
An ablation study with varying $n_k$ can be found in the experiments section.

% However, this similarity measure function might not be sufficient to model the complex source image feature distributions.
% Hence, we incorporate more elaborated version of $h(\cdot)$, which is defined as
% \begin{equation}
%     \tilde{h}(d_i^j) = \tilde{s}_i^j = \sum_{k=1}^{n_k} e^{-\lambda_k d_i^j},
% \end{equation}
% where $n_k$ is the number of basis functions.
% Note here that the sum of rank-preserving functions is also rank-preserving.
% An ablation study with varying $n_k$ can be found in the experiments section.

\subsection{Weighted element-wise means and variances}
We aggregate the features based on the weighted statistical distributions of element-wise means and variances.
For the $k$-th similarity distribution, the weight is derived from the measured similarity as
\begin{equation}
    w_{i,j}^k = s_{i,j}^k/\sum_j s_{i,j}^k.
\end{equation}
Thereafter, the weighted element-wise mean $m_i^k$ and variance $v_i^k$ are calculated as
\begin{equation}
    \mb{m}_i^k[l] = \sum_j w_{i,j}^k \mb{f}_j[l],~\mb{v}_i^k[l] = \sum_j w_{i,j}^k(\mb{f}_j[l]-\mb{m}_i^k[l])^2,
\end{equation}
where $\mb{a}[l]$ is the $l$-th element of $\mb{a}$.

The aggregated feature set $\{\mb{m}_i^k, \mb{v}_i^k|k=1,\cdots,n_k\}$ is concatenated with the $i$-th source image feature $\mb{f}_i$ to be utilized as an input for a neural network which estimates the volume densities and colors of 3D points.

\section{Proposed framework}
\label{sec:framework}
\subsection{Network design}
\begin{figure*}[t]
    \centering
    \includegraphics[height=4.7cm]{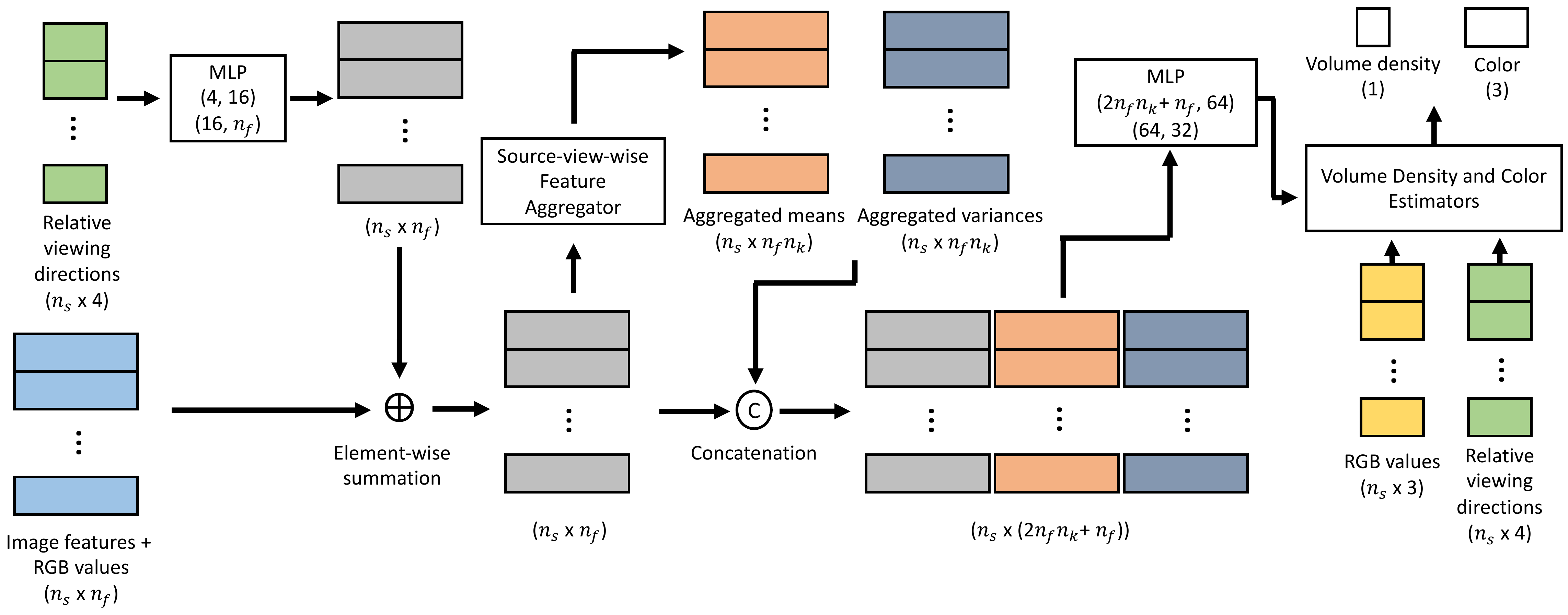}
    \vspace{-2mm}
    \caption{An overview of the proposed network structure incorporating the proposed source-view-wise aggregated features. Here, the feature extraction network is not visualized for simplicity. The inputs of the network are source image features, source image RGB values, and relative viewing direction features, and the network infers the volume density and color of the given 3D point. For each MLP block, ($u$, $v$) represents the linear layer of which the input feature dimension is $u$ and the output feature dimension is $v$. We utilize ELU as the activation function. Here, $n_s$, $n_k$, and $n_f$ are the number of source images, similarity measure functions, and the feature dimensions, respectively.}
    \label{fig:structure}
\end{figure*}

An overview of the proposed network structure incorporating the source-view-wise aggregation features is visualized in Figure \ref{fig:structure}.
We design the network based on the framework of \cite{wang:ibrnet:cvpr2021} as it shows the state-of-the-art performance.
A thing to note here is that the proposed feature aggregation method might be utilized with other multi-view image-based rendering methods which utilize global aggregation features \cite{chen:mvsnerf:arxiv2021, yu:pixelnerf:cvpr2021}.

For the image feature extraction network, we use the same network structure used in \cite{wang:ibrnet:cvpr2021} with the output feature dimension is $16$.
After the source images are encoded with the feature extraction network, the features and RGB values at the projected pixels are fed into the network along with the relative viewing direction features.
Here, the relative viewing direction features model the differences between the target camera viewpoint and the source camera viewpoints.
The relative viewing direction features are fed into a MLP to match the feature dimension of the features with that of the image features.
The dimension-matched relative viewing direction features are added to the image features, and the features are fed into the source-view-wise feature aggregator introduced in Section \ref{sec:proposed}.
The feature aggregator outputs aggregated feature set, which is concatenated with the input feature of the aggregator.
The concatenated features are then fed into a MLP, and the output of the MLP is fed into the volume density and color estimator along with the RGB values and relative viewing direction features.
The structure of the volume density and color estimator is the same as the latter part of the IBRNet structure \cite{wang:ibrnet:cvpr2021}.
For more detailed structures, please refer to \cite{wang:ibrnet:cvpr2021}.

\subsection{Loss function}
Like previous volume-rendering-based methods \cite{mildenhall:nerf:eccv2020, yu:pixelnerf:cvpr2021, wang:ibrnet:cvpr2021}, the whole network is trained in an end-to-end manner leveraging the differentiable volume rendering scheme.
The rendering loss is defined as
\begin{equation}
    L_{\textrm{render}} = \sum_{r\in R} \|\hat{C}(r)-C(r)\|_F^2,
\end{equation}
where $\hat{C}(r)$ and $C(r)$ are the estimated and the ground truth RGB values, respectively. $R$ is the sampled ray batch in a training iteration.

\subsection{Implementation details}
We incorporate all the techniques proposed in \cite{wang:ibrnet:cvpr2021} like neighboring view selection scheme, ray transformer, and temporal visual consistency improvement scheme.
For the effective point sampling on a ray, we adopt the hierarchical sampling scheme following the general practices in \cite{mildenhall:nerf:eccv2020, wang:ibrnet:cvpr2021}.
Hence, we train two networks simultaneously for the coarse and fine point samplings.
In the coarse point sampling, we uniformly sample $n_t=64$ points.
After that, in the fine point sampling, we conduct importance sampling based on the estimated volume densities from the coarse network.
As a result, we additionally sample $64$ points, resulting in $n_t=128$ for the fine network.
To build our framework, we use PyTorch \cite{NEURIPS2019_9015}.
For optimization, we use Adam \cite{kingma2017adam} optimizer. The initial learning rate is set to $1\times 10^{-3}$ for the feature extraction network, and $5\times 10^{-4}$ for the other part of the network.
The learning rate is decreased by a factor of $0.5$ for every $50$k iteration.
The whole training ends at $250$k iterations.
The network is trained on four V100 GPUs with a batch size of $2000$ rays, which takes about a day to finish.
For the network configuration, we use $n_f=19$, $n_k=5$, $n_s=10$, unless stated otherwise.
For the finetuning, we finetune the whole network with a lower learning rate of $5\times 10^{-4}$ and $2\times 10^{-4}$ for the feature extraction network and the other part of the network, respectively.
Here, the learning rate is decreased by a factor of $0.5$ for every $20$k iteration.
The network is finetuned on two V100 GPUs with a batch size of $1000$ rays.
The finetuning process ends at $60$k iterations, which takes about six hours to finish.

\subsection{Comparison with other methods}
\noindent
\textbf{Comparison with GRF.} Though GRF \cite{trevithick2021grf} follows the same motivation which handles occlusion problems, its approach is different from our method.
While GRF uses attention to aggregate source 2D features into globally aggregated 3D point feature, our method uses MLP to integrate both local and global information to infer the volume density and color of a 3D point.
Because GRF does not integrate local and global information, it can be ineffective to infer surfaces which leads to inferior performance in occlusion problems.
A quantitative comparison result can be found in the experiments section.\\
\textbf{Comparison with self-attention.} The proposed method may seem similar to the self-attention mechanism \cite{vaswani:selfattention:neurips2017} in that the element-wise similarities are utilized.
However, the proposed feature aggregation method is different from the self-attention scheme in two ways: (i) We incorporate $l_2$-norm as the distance metric, which is commonly used in the multi-view image-based rendering approach.
This might enable us to use the proposed method as a universal technique.
(ii) The number of additional parameters to utilize the self-similarities is much fewer than that of the self-attention mechanism.
For example, in GRF, AttSets needs about 1000k additional parameters, and Slot Attention needs about 300k additional parameters.
However, the proposed method just needs $n_k=5$ additional parameters.

\section{Experiments}
\label{sec:performance}
In this section, we show experimental results on two evaluation protocols.
On the first protocol of ``pretrained", the model is trained on training scenes and tested on unobserved scenes without any further optimization.
On the second protocol of ``finetuned", the model trained with the first protocol is finetuned using the training images of given test scenes before rendering the novel view images of the test scenes.

\begin{figure*}[t]
    \centering
    \includegraphics[height=15cm]{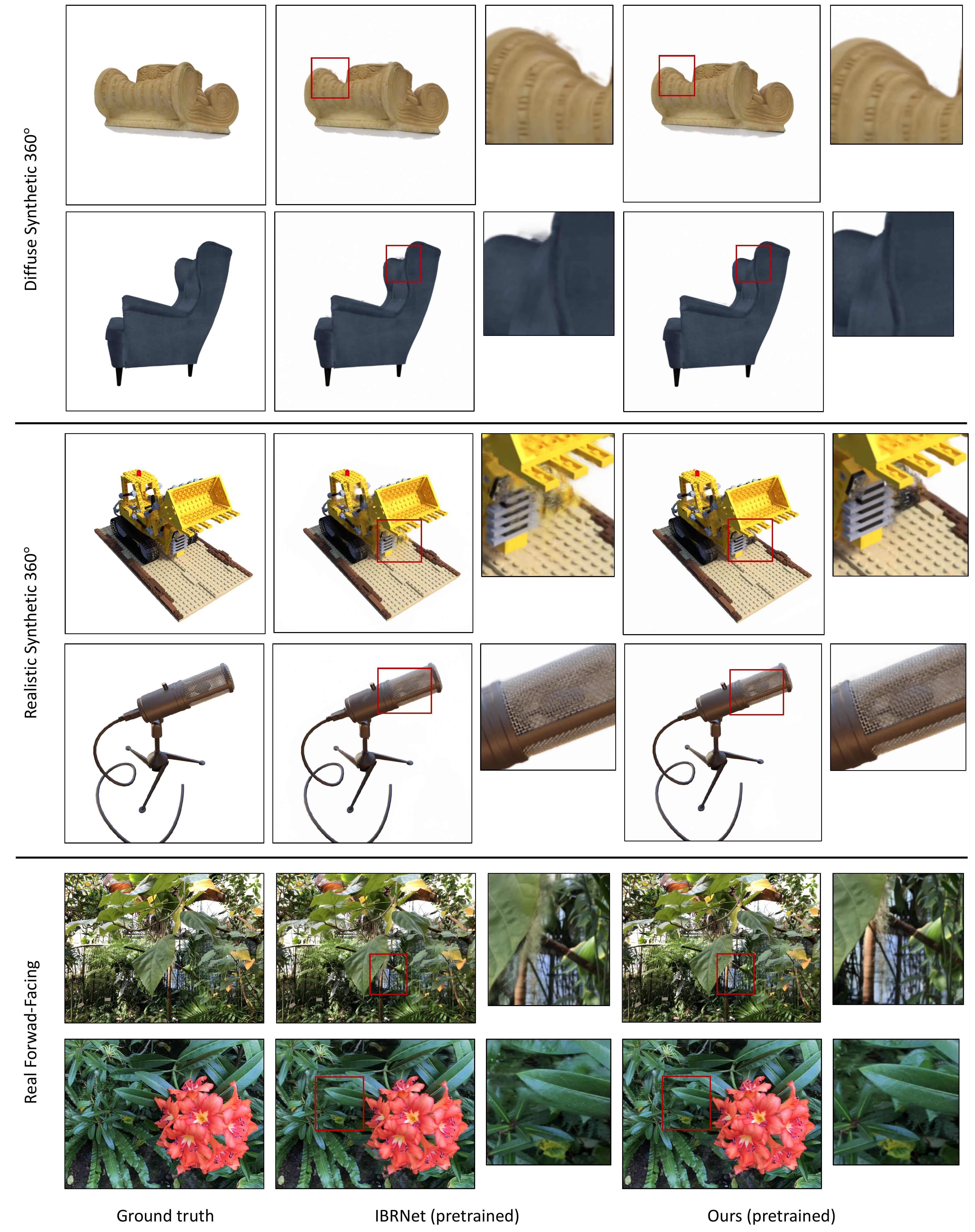}
    \caption{Qualitative comparison results on the benchmark datasets with the ``pretrained'' protocol. The proposed method shows more accurate results compared to the other method.}
    \label{fig:qualitative}
\end{figure*}

\subsection{Dataset}
For training our model based on the ``pretrained" protocol, we incorporate five public datasets: object-centrically rendered scenes using a subset of the Google Scanned Objects dataset \cite{googlescanned} which is provided by \cite{wang:ibrnet:cvpr2021}, the RealEstate10K dataset \cite{zhou:stereomagni:siggraph2018}, the Spaces dataset \cite{flynn:deepview:cvpr2019}, the LLFF dataset \cite{mildenhall:llff:acm2019}, and the IBRNet collected dataset \cite{wang:ibrnet:cvpr2021}.
The training datasets consist of various synthetic and real scenes, which facilitates us to train the generalizable model.

To evaluate the proposed model, we utilize three datasets: the Diffuse Synthetic 360$^{\circ}$ dataset \cite{sitzmann:deepvoxels:cvpr2019}, the Realistic Synthetic 360$^{\circ}$ dataset \cite{mildenhall:nerf:eccv2020}, and the Forward-Facing \cite{mildenhall:llff:acm2019} dataset.
The test datasets consist of real scene \cite{mildenhall:llff:acm2019} and synthetic scenes \cite{sitzmann:deepvoxels:cvpr2019, mildenhall:nerf:eccv2020}, which allows us to evaluate the proposed method in various situations.
Following the practice in \cite{wang:ibrnet:cvpr2021}, we evaluate the proposed model on the sub-sampled images with a factor of $10$, $8$, and $8$ for the Diffuse Synthetic 360$^{\circ}$ dataset, the Realistic Synthetic 360$^{\circ}$ dataset, and the Forward-Facing dataset, respectively.
For all the datasets, we utilize the camera parameters and the near and far bounds estimated using COLMAP \cite{schonberger:colmap:cvpr2016}.

\subsection{Quantitative results}

In this section, we show the quantitative comparison results.
For the evaluation metric, we incorporate peak signal to noise ratio (PSNR), structural similarity index measure (SSIM) \cite{ssim}, and leaned perceptual image patch similarity (LPIPS) \cite{lpips}, following the practices in \cite{wang:ibrnet:cvpr2021}.
We first compare the performance of the proposed method based on the ``pretrained" protocol with state-of-the-art methods of LLFF \cite{mildenhall:llff:acm2019} and IBRNet \cite{wang:ibrnet:cvpr2021}.
The comparison results are summarized in the top part of Table \ref{tab:main_result}.
From the results, we can demonstrate that the proposed method shows the state-of-the-art performance in most cases.
On the Diffuse Synthetic 360$^{\circ}$ dataset, the proposed method shows comparable performances with IBRNet on the SSIM and LPIPS measures.
We conjecture that the proposed method has little room for improvement on these cases as the Diffuse Synthetic 360$^{\circ}$ dataset consists of simple scenes.
However, we note that the proposed method improves the PSNR measure by a large margin.
In other cases, which are relatively hard, we can see that the proposed method shows the state-of-the-art performances.

We also report the performance of the proposed method based on the ``finetuned" protocol.
We compare the performance of the proposed method with various methods of SRN \cite{sitzmann:srn:neurips2019}, NV \cite{lombardi:neuralvolumes:arxiv2019}, NeRF \cite{mildenhall:nerf:eccv2020}, and IBRNet \cite{wang:ibrnet:cvpr2021}.
The comparison results are summarized in the bottom part of Table \ref{tab:main_result}.
The results demonstrate that the proposed method shows better or competitive performances compared with one of the state-of-the-art methods, IBRNet.

In addition, we report the comparison result with NeRF and GRF \cite{trevithick2021grf} in Table \ref{tab:main_result_grf} based on the both of the protocols.
For the fair comparison, we have followed the practices of \cite{trevithick2021grf}.
Specifically, the network is trained on Chair, Mic, Ship, and Hotdog scenes of the Realistic Synthetic 360$^{\circ}$ dataset, and tested on Drums, Lego, Materials, and Ficus scenes of the Realistic Synthetic 360$^{\circ}$ dataset.
In this case, we use four neighboring source view images, i.e., $n_s$=4.
From the results, we can see that the proposed method shows much better performance than GRF.

\begin{table*}[t]
\centering
\caption{Performance comparison results on the several benchmark datasets including real and synthetic scenes. For the protocol, ``P'' means the ``pretrained'' protocol, and ``F'' means the ``finetuned'' protocol. The performances of the other methods are quoted from \cite{wang:ibrnet:cvpr2021}. In each category, the best results are in \tb{bold}, and the second-best are \tu{underlined}.}
\vspace{-2mm}
\resizebox{0.92\textwidth}{!}{%
\begin{tabular}{l|c|ccc|ccc|ccc} \toprule
\multirow{2}{*}{Method} & \multirow{2}{*}{Protocol} & \multicolumn{3}{|c}{Diffuse Synthetic 360$^{\circ}$ \cite{sitzmann:deepvoxels:cvpr2019}} & \multicolumn{3}{|c}{Realistic Synthetic 360$^{\circ}$ \cite{mildenhall:nerf:eccv2020}} & \multicolumn{3}{|c}{Real Forward-Facing \cite{mildenhall:llff:acm2019}} \\
\cline{3-11} & & PSNR$\uparrow$ & SSIM$\uparrow$ & LPIPS$\downarrow$ & PSNR$\uparrow$ & SSIM$\uparrow$ & LPIPS$\downarrow$ & PSNR$\uparrow$ & SSIM$\uparrow$ & LPIPS$\downarrow$  \\
\toprule
LLFF \cite{mildenhall:llff:acm2019} & P & 34.38 & 0.985 & 0.048 & 24.88 & 0.911 & 0.114 & 24.13 & 0.798 & 0.212\\
IBRNet \cite{wang:ibrnet:cvpr2021} & P & \tu{37.17} & \tb{0.990} & \tb{0.017} & \tu{25.49} & \tu{0.916} & \tu{0.100} & \tu{25.13} & \tu{0.817} & \tu{0.205} \\
Ours & P & \tb{37.45} & \tb{0.990} & \tb{0.017} & \tb{27.13} & \tb{0.930} & \tb{0.083} & \tb{25.47} & \tb{0.826} & \tb{0.196} \\
\hline
SRN \cite{sitzmann:srn:neurips2019} &  F & 33.20 & 0.963 & 0.073 & 22.26 & 0.846 & 0.170 & 22.84 & 0.668 & 0.378 \\
NV \cite{lombardi:neuralvolumes:arxiv2019} & F & 29.62 & 0.929 & 0.099 & 26.05 & 0.893 & 0.160 & - & - & - \\
NeRF \cite{mildenhall:nerf:eccv2020} & F & 40.15 & 0.991 & 0.023 & \tb{31.01} & \tu{0.947} & 0.081 & 26.50 & 0.811 & 0.250 \\
IBRNet \cite{wang:ibrnet:cvpr2021} & F & \tu{42.93} & \tb{0.997} & \tu{0.009} & 28.14 & 0.942 & \tu{0.072} & \tb{26.73} & \tb{0.851} & \tb{0.175} \\
%\hline
Ours & F & \tb{43.21} & \tb{0.997} & \tb{0.008} & \tu{30.89} & \tb{0.956} & \tb{0.051} & \tu{26.66} & \tu{0.849} & \tb{0.175} \\
\bottomrule
\end{tabular}%
}

\label{tab:main_result}
\end{table*}

\begin{table*}[t]
\centering
\caption{Performance comparison results on the Realistic Synthetic 360$^{\circ}$ dataset. The performance of the other methods are quoted from \cite{trevithick2021grf}. In each category, the best results are in \tb{bold}, and the second-best are \tu{underlined}.}
\vspace{-2mm}
\resizebox{0.95\textwidth}{!}{%
\begin{tabular}{l|ccc|ccc|ccc|ccc} \toprule
\multirow{2}{*}{} &  \multicolumn{3}{|c}{Pretrained} & \multicolumn{3}{|c}{Finetuned (100 iters)} &  \multicolumn{3}{|c}{Finetuned (1k iters)} & \multicolumn{3}{|c}{Finetuned (10k iters)}  \\
\cline{2-13} & PSNR$\uparrow$ & SSIM$\uparrow$ & LPIPS$\downarrow$ & PSNR$\uparrow$ & SSIM$\uparrow$ & LPIPS$\downarrow$ & PSNR$\uparrow$ & SSIM$\uparrow$ & LPIPS$\downarrow$ & PSNR$\uparrow$ & SSIM$\uparrow$ & LPIPS$\downarrow$ \\
\toprule
NeRF & - & - & - & 15.15 & 0.752 & 0.359 & 19.81 & 0.809 & 0.228 & 23.35 & 0.875 & 0.137 \\
GRF  & \tu{13.62} & \tu{0.763} & \tu{0.246} & \tu{19.69} & \tu{0.835} & \tu{0.169} & \tu{22.00} & \tu{0.876} & \tu{0.128} & \tu{25.10} & \tu{0.916} & \tu{0.089} \\
Ours  & \tb{22.37} & \tb{0.904} & \tb{0.096} & \tb{25.58} & \tb{0.928} & \tb{0.076} & \tb{26.86} & \tb{0.941} & \tb{0.061} & \tb{27.95} & \tb{0.950} & \tb{0.053} \\
\bottomrule
\end{tabular}%
}

\label{tab:main_result_grf}
\end{table*}

\subsection{Qualitative results}
We provide qualitative comparison results on the ``pretrained'' protocol with IBRNet \cite{wang:ibrnet:cvpr2021} which is one of the state-of-the-art methods.
Note that the qualitative comparison results on the ``finetuned'' protocol can be found in Appendix.
The comparison results on diverse scenes are visualized in Figure \ref{fig:qualitative}.
We can see that the proposed method synthesizes the novel view image more accurately than IBRNet in all the cases.
Specifically, in the Diffuse Synthetic 360$^{\circ}$ dataset, the proposed method shows better results in the edge regions.
The results of IBRNet have some artifacts, which is not the case for the proposed method.
The source image feature set in the edge region is easy to include outlier features of the background, which is handled by incorporating the proposed aggregation features in the proposed method.
In the case of the Realistic Synthetic 360$^{\circ}$ dataset, the proposed method synthesizes the image more accurately in the region which is easy to be self-occluded.
The proposed method successfully synthesizes the poclain belt and the inside structure of the microphone.
In the case of the Real Forward-Facing dataset, the proposed method shows more accurate results in the region which is easy to be self-occluded (the first example) and in the edge region (the second example).
These results demonstrate that the proposed method facilitates us to utilize robust aggregation features in situations where some outliers are included in the source image feature set.

\subsection{Ablation study}
\label{subsection:ablation}
%In order to provide extra analysis of the proposed approach, we evaluate our method with various settings.\\
\noindent
\textbf{Various values of $n_k$.} We report the ablative results with respect to various values of $n_k$ in Table \ref{tab:ablation}.
We can see that incorporating the proposed source-view-wise aggregation features improves the baseline \cite{wang:ibrnet:cvpr2021} performance on the Realistic Synthetic 360$^{\circ}$ and Real Forward-Facing datasets even when we utilize only one learnable similarity measure function.
From the results, we can demonstrate the effectiveness of the proposed scheme over the global feature aggregation method.
In the case of the Diffuse Synthetic 360$^{\circ}$ dataset, the proposed method shows comparable performance with the baseline.
We speculate that this comes from the fact that the Diffuse Synthetic 360$^{\circ}$ dataset consists of simple scenes, resulting in little room for improvement.
On the one hand, we can also validate the robustness of the proposed framework with respect to the choice of $n_k$.\\
%Note that the performances are gradually increased as we incorporate a larger number of the proposed similarity measure functions.
%We conjecture that the improved performance comes from the increased feature distribution modeling capacity of the proposed feature aggregator.\\
\textbf{Choice of the distance metric.} In order to validate the benefit of incorporating $l_2$-norm as the distance metric, we check the performance of the proposed scheme in case of utilizing the cosine distance as the distance metric.
Note that the $l_2$-norm is commonly used in the multi-view image-based rendering approach \cite{wang:ibrnet:cvpr2021, chen:mvsnerf:arxiv2021}, and the cosine distance is widely utilized in the self-attention mechanism \cite{vaswani:selfattention:neurips2017}.
To this end, we measure the distance between two features $f_i$, $f_j$ as $d_{i,j}=1-f_i\cdot f_j/\|f_i\|_2\|f_j\|_2$.
Here, $\cdot$ means the dot product operation and $\|\cdot\|_2$ is the $l_2$-norm.
The result is summarized in Table \ref{tab:ablation_2} (Exp1).
We can see that incorporating the cosine distance as the distance metric degrades the performance in most of the cases.\\
\textbf{Choice of the similarity measure function.} We conduct an ablative study with the mapping functions in the form of 1/(1+$\lambda_k d_{i,j}$).
The result is summarized in Table \ref{tab:ablation_2} (Exp2).
We can see that incorporating this form slightly decreases the performance.\\
%Note that more complex loss function design is needed to incorporate MLP as the mapping function to preserve the rank constraint.
%Also, we need more learning parameters.
%Hence, we argue that this could not be a favorable choice.\\
\textbf{Learnability of $\lambda$.} To check the efficacy of the learnability of the similarity measure functions, we report the performance with evenly distributed fixed lambdas ($\lambda$ = 0.05, 1.2875, 2.525, 3.7625, 5).
The result is summarized in Table \ref{tab:ablation_2} (Exp3).
We can see the decreased performance, which validates the importance of the learnability.

\begin{table*}[t]
\centering
\caption{Ablation results with respect to various values of $n_k$ on the benchmark datasets. For the protocol, ``P'' means the ``pretrained'' protocol. The best results are in \tb{bold}, and the second-best are \tu{underlined}.}
\vspace{-2mm}
\resizebox{0.95\textwidth}{!}{%
\begin{tabular}{l|c|ccc|ccc|ccc} \toprule
\multirow{2}{*}{Method} & \multirow{2}{*}{Protocol} & \multicolumn{3}{|c}{Diffuse Synthetic 360$^{\circ}$ \cite{sitzmann:deepvoxels:cvpr2019}} & \multicolumn{3}{|c}{Realistic Synthetic 360$^{\circ}$ \cite{mildenhall:nerf:eccv2020}} & \multicolumn{3}{|c}{Real Forward-Facing \cite{mildenhall:llff:acm2019}} \\
\cline{3-11} & & PSNR$\uparrow$ & SSIM$\uparrow$ & LPIPS$\downarrow$ & PSNR$\uparrow$ & SSIM$\uparrow$ & LPIPS$\downarrow$ & PSNR$\uparrow$ & SSIM$\uparrow$ & LPIPS$\downarrow$  \\
\toprule
Baseline \cite{wang:ibrnet:cvpr2021} & P & 37.17 & \tb{0.990} & \tb{0.017} & 25.49 & 0.916 & 0.100 & 25.13 & 0.817 & 0.205 \\
$n_k=1$ & P & 37.01 & 0.989 & 0.020 & 26.81 & 0.927 & 0.085 & 25.31 & 0.821 & 0.201 \\
$n_k=2$ & P & 37.20 & 0.989 & 0.019 & 26.96 & 0.928 & 0.083 & 25.37 & 0.823 & 0.201 \\
$n_k=3$ & P & \tu{37.28} & 0.989 & 0.018 & \tb{27.24} & \tb{0.931} & \tb{0.082} & 25.36 & 0.822 & 0.200 \\
$n_k=4$ & P & \tu{37.28} & 0.989 & 0.018 & \tu{27.14} & 0.929 & \tb{0.082} & \tb{25.47} & \tu{0.825} & \tu{0.198} \\
%\hline
Ours ($n_k=5$) & P & \tb{37.45} & \tb{0.990} & \tb{0.017} & 27.13 & \tu{0.930} & 0.083 & \tb{25.47} & \tb{0.826} & \tb{0.196} \\
\bottomrule
\end{tabular}%
}

\label{tab:ablation}
\end{table*}

\begin{table*}[t]
\centering
\caption{Ablation results with respect to various settings on the benchmark datasets. For the protocol, ``P'' means the ``pretrained'' protocol. The best results are in \tb{bold}.}
\vspace{-2mm}
\resizebox{0.99\textwidth}{!}{%
\begin{tabular}{l|c|ccc|ccc|ccc} \toprule
\multirow{2}{*}{Method} & \multirow{2}{*}{Protocol} & \multicolumn{3}{|c}{Diffuse Synthetic 360$^{\circ}$ \cite{sitzmann:deepvoxels:cvpr2019}} & \multicolumn{3}{|c}{Realistic Synthetic 360$^{\circ}$ \cite{mildenhall:nerf:eccv2020}} & \multicolumn{3}{|c}{Real Forward-Facing \cite{mildenhall:llff:acm2019}} \\
\cline{3-11} & & PSNR$\uparrow$ & SSIM$\uparrow$ & LPIPS$\downarrow$ & PSNR$\uparrow$ & SSIM$\uparrow$ & LPIPS$\downarrow$ & PSNR$\uparrow$ & SSIM$\uparrow$ & LPIPS$\downarrow$  \\
\toprule
Exp1 (cosine distance) & P & 37.26 & 0.989 & 0.018 & 27.09 & \tb{0.930} & 0.082 & 25.39 & 0.824 & 0.199 \\
%Exp2 (harsh; mean+var aggre) & P & 33.99 & 0.979 & 0.034 & 22.82 & 0.877 & 0.145 & 23.20 & 0.765 & 0.264 \\
%Exp3 (harsh; our aggre) & P & 34.72 & 0.982 & 0.028 & 23.61 & 0.890 & 0.130 & 23.50 & 0.774 & 0.256 \\
%Exp4 (mean aggre) & P & 36.94 & 0.988 & 0.021 & 26.66 & 0.924 & 0.091 & 25.10 & 0.816 & 0.207 \\
%Exp5 (mean+var aggre) & P & 37.03 & 0.988 & 0.019 & 26.72 & 0.925 & 0.090 & 25.16 & 0.817 & 0.204 \\
%Exp5 (atten aggre) & P & 36.77 & 0.988 & 0.020 & 26.76 & 0.924 & 0.088 & 25.15 & 0.818 & 0.204 \\
Exp2 (1/$\lambda d$) & P & 37.42 & 0.989 & 0.018 & 27.10 & \tb{0.930} & \tb{0.080} & 25.33 & 0.823 & 0.200 \\
Exp3 (fixed $\lambda$) & P & 37.15 & 0.989 & 0.019 & 26.93 & 0.928 & 0.083 & 25.39 & 0.824 & 0.198 \\
\hline
Ours & P &  \tb{37.45} & \tb{0.990} & \tb{0.017} & \tb{27.13} & \tb{0.930} & 0.083 & \tb{25.47} & \tb{0.826} & \tb{0.196} \\
\bottomrule
\end{tabular}%
}
\label{tab:ablation_2}
\end{table*}

\section{Conclusions}
\label{sec:conclusion}
In this paper, we have proposed a novel source-view-wise feature aggregation method for the multi-view image-based rendering problem.
Unlike the previous methods that incorporate equally-weighted aggregation features to figure out surfaces in the estimation process, the proposed method uses a set of locally aggregated features to facilitate us to leverage the local structures in the feature set.
In experiments, incorporating the proposed features improves the performance by a large margin, resulting in the state-of-the-art performances.
This demonstrates the effectiveness of the proposed scheme.
A thing to note here is that the proposed method can be applied to other baseline methods other than the one we used in the implementation.
The proposed method could be more effective when used with more powerful baselines, which is left as future work.

\clearpage
\bibliographystyle{splncs04}
\bibliography{ref}

\begin{thebibliography}{10}
\providecommand{\url}[1]{\texttt{#1}}
\providecommand{\urlprefix}{URL }
\providecommand{\doi}[1]{https://doi.org/#1}

\bibitem{barron:mipnerf:iccv2021}
Barron, J.T., Mildenhall, B., Tancik, M., Hedman, P., Martin-Brualla, R.,
  Srinivasan, P.P.: {Mip-NeRF: A Multiscale Representation for Anti-Aliasing
  Neural Radiance Fields}. In: Proc. IEEE Int'l Conf. Computer Vision (October
  2021)

\bibitem{chen:mvsnerf:arxiv2021}
Chen, A., Xu, Z., Zhao, F., Zhang, X., Xiang, F., Yu, J., Su, H.: Mvsnerf: Fast
  generalizable radiance field reconstruction from multi-view stereo. arXiv
  preprint arXiv:2103.15595  (2021)

\bibitem{chibane:srf:cvpr2021}
Chibane, J., Bansal, A., Lazova, V., Pons-Moll, G.: Stereo radiance fields
  (srf): Learning view synthesis for sparse views of novel scenes. In: Proc.
  IEEE Conf. Computer Vision and Pattern Recognition (June 2021)

\bibitem{flynn:deepview:cvpr2019}
Flynn, J., Broxton, M., Debevec, P., DuVall, M., Fyffe, G., Overbeck, R.,
  Snavely, N., Tucker, R.: {DeepView: View synthesis with learned gradient
  descent}. In: Proc. IEEE Conf. Computer Vision and Pattern Recognition (June
  2019)

\bibitem{flynn:deepstereo:cvpr2016}
Flynn, J., Neulander, I., Philbin, J., Snavely, N.: {Deepstereo: Learning to
  Predict New Views from the World's Imagery}. In: Proc. IEEE Conf. Computer
  Vision and Pattern Recognition (June 2016)

\bibitem{genova:ldif:cvpr2020}
Genova, K., Cole, F., Sud, A., Sarna, A., Funkhouser, T.: {Local Deep Implicit
  Functions for 3D Shape}. In: Proc. IEEE Conf. Computer Vision and Pattern
  Recognition (June 2020)

\bibitem{googlescanned}
{Google Research}: Google scanned objects.
  \url{https://app.ignitionrobotics.org/GoogleResearch/fuel/collections/GoogleScannedObjects}

\bibitem{jain:diernerf:iccv2021}
Jain, A., Tancik, M., Abbeel, P.: {Putting NeRF on a Diet: Semantically
  Consistent Few-Shot View Synthesis}. In: Proc. IEEE Int'l Conf. Computer
  Vision (October 2021)

\bibitem{jiang:localimplicitgrid:cvpr2020}
Jiang, C., Sud, A., Makadia, A., Huang, J., Nie{\ss}ner, M., Funkhouser, T.,
  et~al.: {Local Implicit Grid Representations for 3D Scenes}. In: Proc. IEEE
  Conf. Computer Vision and Pattern Recognition (June 2020)

\bibitem{kajiya:volumnrender:1984}
Kajiya, J.T., Von~Herzen, B.P.: Ray tracing volume densities. ACM SIGGRAPH
  computer graphics  \textbf{18}(3),  165--174 (1984)

\bibitem{kalantari:learning:acmtog2016}
Kalantari, N.K., Wang, T.C., Ramamoorthi, R.: {Learning-Based View Synthesis
  for Light Field Cameras}. ACM TOG  (2016)

\bibitem{kar:learning:neurips2017}
Kar, A., H{\"a}ne, C., Malik, J.: {Learning a Multi-View Stereo Machine}. In:
  Advances in neural information processing systems (2017)

\bibitem{kingma2017adam}
Kingma, D.P., Ba, J.: Adam: A method for stochastic optimization (2017)

\bibitem{li:neuralff:cvpr2021}
Li, Z., Niklaus, S., Snavely, N., Wang, O.: {Neural Scene Flow Fields for
  Space-Time View Synthesis of Dynamic Scenes}. In: Proc. IEEE Conf. Computer
  Vision and Pattern Recognition (June 2021)

\bibitem{li:crowdsampling:eccv2020}
Li, Z., Xian, W., Davis, A., Snavely, N.: {Crowdsampling the Plenoptic
  Function}. In: Proc. European Conference on Computer Vision (August 2020)

\bibitem{lin:barf:iccv2021}
Lin, C.H., Ma, W.C., Torralba, A., Lucey, S.: {BARF: Bundle-Adjusting Neural
  Radiance Fields}. In: Proc. IEEE Int'l Conf. Computer Vision (October 2021)

\bibitem{lindell:autoint:cvpr2021}
Lindell, D.B., Martel, J.N., Wetzstein, G.: {AutoInt: Automatic Integration for
  Fast Neural Volume Rendering}. In: Proc. IEEE Conf. Computer Vision and
  Pattern Recognition (June 2021)

\bibitem{liu:nsvf:neurips2020}
Liu, L., Gu, J., Lin, K.Z., Chua, T.S., Theobalt, C.: {Neural Sparse Voxel
  Fields}. In: Advances in neural information processing systems (2020)

\bibitem{liu:dist:cvpr2020}
Liu, S., Zhang, Y., Peng, S., Shi, B., Pollefeys, M., Cui, Z.: {Dist: Rendering
  Deep Implicit Signed Distance Function with Differentiable Sphere Tracing}.
  In: Proc. IEEE Conf. Computer Vision and Pattern Recognition (June 2020)

\bibitem{liu:editnerf:iccv2021}
Liu, S., Zhang, X., Zhang, Z., Zhang, R., Zhu, J.Y., Russell, B.: Editing
  conditional radiance fields. In: Proc. IEEE Int'l Conf. Computer Vision
  (October 2021)

\bibitem{lombardi:neuralvolumes:arxiv2019}
Lombardi, S., Simon, T., Saragih, J., Schwartz, G., Lehrmann, A., Sheikh, Y.:
  {Neural Volumes: Learning Dynamic Renderable Volumes from Images}. ACM TOG
  (2019)

\bibitem{martin:nerfinwild:cvpr2021}
Martin-Brualla, R., Radwan, N., Sajjadi, M.S., Barron, J.T., Dosovitskiy, A.,
  Duckworth, D.: {NeRF in the Wild: Neural Radiance Fields for Unconstrained
  Photo Collections}. In: Proc. IEEE Conf. Computer Vision and Pattern
  Recognition (June 2021)

\bibitem{mescheder:occupancynetwork:cvpr2019}
Mescheder, L., Oechsle, M., Niemeyer, M., Nowozin, S., Geiger, A.: {Occupancy
  Networks: Learning 3D Reconstruction in Function Space}. In: Proc. IEEE Conf.
  Computer Vision and Pattern Recognition (June 2019)

\bibitem{mildenhall:llff:acm2019}
Mildenhall, B., Srinivasan, P.P., Ortiz-Cayon, R., Kalantari, N.K.,
  Ramamoorthi, R., Ng, R., Kar, A.: {Local Light Field Fusion: Practical View
  Synthesis with Prescriptive Sampling Guidelines}. ACM TOG  (2019)

\bibitem{mildenhall:nerf:eccv2020}
Mildenhall, B., Srinivasan, P.P., Tancik, M., Barron, J.T., Ramamoorthi, R.,
  Ng, R.: Nerf: Representing scenes as neural radiance fields for view
  synthesis. In: Proc. European Conference on Computer Vision (August 2020)

\bibitem{niemeyer:dvr:cvpr2020}
Niemeyer, M., Mescheder, L., Oechsle, M., Geiger, A.: {Differentiable
  Volumetric Rendering: Learning Implicit 3D Representations without 3D
  Supervision}. In: Proc. IEEE Conf. Computer Vision and Pattern Recognition
  (June 2020)

\bibitem{park:deepsdf:cvpr2019}
Park, J.J., Florence, P., Straub, J., Newcombe, R., Lovegrove, S.: {DeepSDF:
  Learning Continuous Signed Distance Functions for Shape Representation}. In:
  Proc. IEEE Conf. Computer Vision and Pattern Recognition (June 2019)

\bibitem{NEURIPS2019_9015}
Paszke, A., Gross, S., Massa, F., Lerer, A., Bradbury, J., Chanan, G., Killeen,
  T., Lin, Z., Gimelshein, N., Antiga, L., Desmaison, A., Kopf, A., Yang, E.,
  DeVito, Z., Raison, M., Tejani, A., Chilamkurthy, S., Steiner, B., Fang, L.,
  Bai, J., Chintala, S.: Pytorch: An imperative style, high-performance deep
  learning library. In: Wallach, H., Larochelle, H., Beygelzimer, A.,
  d\textquotesingle Alch\'{e}-Buc, F., Fox, E., Garnett, R. (eds.) Advances in
  Neural Information Processing Systems 32, pp. 8024--8035. Curran Associates,
  Inc. (2019)

\bibitem{penner:soft3d:acm2017}
Penner, E., Zhang, L.: {Soft 3D Reconstruction for View Synthesis}. SIGGRAPH
  Asia  (2017)

\bibitem{rahaman2019spectral}
Rahaman, N., Baratin, A., Arpit, D., Draxler, F., Lin, M., Hamprecht, F.,
  Bengio, Y., Courville, A.: On the spectral bias of neural networks. In:
  International Conference on Machine Learning. pp. 5301--5310. PMLR (2019)

\bibitem{rebain:derf:cvpr2021}
Rebain, D., Jiang, W., Yazdani, S., Li, K., Yi, K.M., Tagliasacchi, A.: {DeRF:
  Decomposed Radiance Fields}. In: Proc. IEEE Conf. Computer Vision and Pattern
  Recognition (June 2021)

\bibitem{saito:pifu:cvpr2019}
Saito, S., Huang, Z., Natsume, R., Morishima, S., Kanazawa, A., Li, H.: {PIFu:
  Pixel-Aligned Implicit Function for High-Resolution Clothed Human
  Digitization}. In: Proceedings of the IEEE International Conference on
  Computer Vision (October 2019)

\bibitem{scharstein2002taxonomy}
Scharstein, D., Szeliski, R.: A taxonomy and evaluation of dense two-frame
  stereo correspondence algorithms. International journal of computer vision
  \textbf{47}(1),  7--42 (2002)

\bibitem{schonberger:colmap:cvpr2016}
Schonberger, J.L., Frahm, J.M.: Structure-from-motion revisited. In: Proc. IEEE
  Conf. Computer Vision and Pattern Recognition (June 2016)

\bibitem{sitzmann:deepvoxels:cvpr2019}
Sitzmann, V., Thies, J., Heide, F., Nie{\ss}ner, M., Wetzstein, G., Zollhofer,
  M.: {DeepVoxels: Learning Persistent 3D Feature Embeddings}. In: Proc. IEEE
  Conf. Computer Vision and Pattern Recognition (June 2019)

\bibitem{sitzmann:srn:neurips2019}
Sitzmann, V., Zollh{\"o}fer, M., Wetzstein, G.: {Scene Representation Networks:
  Continuous 3D-Structure-Aware Neural Scene Representations}. In: Advances in
  neural information processing systems (2019)

\bibitem{srinivasan:mpiextra:cvpr2019}
Srinivasan, P.P., Tucker, R., Barron, J.T., Ramamoorthi, R., Ng, R., Snavely,
  N.: {Pushing the Boundaries of View Extrapolation with Multiplane Images}.
  In: Proc. IEEE Conf. Computer Vision and Pattern Recognition (June 2019)

\bibitem{szeliski2007image}
Szeliski, R., et~al.: Image alignment and stitching: A tutorial. Foundations
  and Trends{\textregistered} in Computer Graphics and Vision  \textbf{2}(1),
  1--104 (2007)

\bibitem{tretschk:nonrigidnerf:cvpr2021}
Tretschk, E., Tewari, A., Golyanik, V., Zollh{\"o}fer, M., Lassner, C.,
  Theobalt, C.: {Non-Rigid Neural Radiance Fields: Reconstruction and Novel
  View Synthesis of a Dynamic Scene From Monocular Video}. In: Proc. IEEE Conf.
  Computer Vision and Pattern Recognition (June 2021)

\bibitem{trevithick2021grf}
Trevithick, A., Yang, B.: {GRF: Learning a General Radiance Field for 3D
  Representation and Rendering}. In: Proc. IEEE Int'l Conf. Computer Vision
  (October 2021)

\bibitem{tucker:sigleviewmpi:cvpr2020}
Tucker, R., Snavely, N.: {Single-View View Synthesis with Multiplane Images}.
  In: Proc. IEEE Conf. Computer Vision and Pattern Recognition (June 2020)

\bibitem{tulsiani:multi:cvpr2017}
Tulsiani, S., Zhou, T., Efros, A.A., Malik, J.: {Multi-view Supervision for
  Single-view Reconstruction via Differentiable Ray Consistency}. In: Proc.
  IEEE Conf. Computer Vision and Pattern Recognition (July 2017)

\bibitem{vaswani:selfattention:neurips2017}
Vaswani, A., Shazeer, N., Parmar, N., Uszkoreit, J., Jones, L., Gomez, A.N.,
  Kaiser, {\L}., Polosukhin, I.: Attention is all you need. In: Advances in
  neural information processing systems (2017)

\bibitem{wang:ibrnet:cvpr2021}
Wang, Q., Wang, Z., Genova, K., Srinivasan, P.P., Zhou, H., Barron, J.T.,
  Martin-Brualla, R., Snavely, N., Funkhouser, T.: Ibrnet: Learning multi-view
  image-based rendering. In: Proc. IEEE Conf. Computer Vision and Pattern
  Recognition (June 2021)

\bibitem{ssim}
Wang, Z., Bovik, A., Sheikh, H., Simoncelli, E.: Image quality assessment: from
  error visibility to structural similarity. IEEE Transactions on Image
  Processing  \textbf{13}(4),  600--612 (2004). \doi{10.1109/TIP.2003.819861}

\bibitem{yariv:multiview:neurips2020}
Yariv, L., Kasten, Y., Moran, D., Galun, M., Atzmon, M., Basri, R., Lipman, Y.:
  {Multiview Neural Surface Reconstruction by Disentangling Geometry and
  Appearance}. In: Advances in neural information processing systems (2020)

\bibitem{yu:plenoctrees:iccv2021}
Yu, A., Li, R., Tancik, M., Li, H., Ng, R., Kanazawa, A.: {PlenOctrees for
  Real-time Rendering of Neural Radiance Fields}. In: Proc. IEEE Int'l Conf.
  Computer Vision (October 2021)

\bibitem{yu:pixelnerf:cvpr2021}
Yu, A., Ye, V., Tancik, M., Kanazawa, A.: pixelnerf: Neural radiance fields
  from one or few images. In: Proc. IEEE Conf. Computer Vision and Pattern
  Recognition (June 2021)

\bibitem{lpips}
Zhang, R., Isola, P., Efros, A.A., Shechtman, E., Wang, O.: The unreasonable
  effectiveness of deep features as a perceptual metric. In: Proc. IEEE Conf.
  Computer Vision and Pattern Recognition (June 2018)

\bibitem{zhou:stereomagni:siggraph2018}
Zhou, T., Tucker, R., Flynn, J., Fyffe, G., Snavely, N.: {Stereo Magnification:
  Learning view synthesis using multiplane images}. SIGGRAPH  (2018)

\end{thebibliography}

\setcounter{section}{0}
\renewcommand\thesection{\Alph{section}}
\renewcommand\thesubsection{\thesection.\arabic{subsection}}

\clearpage

\section{Appendix}
\subsection{Analysis of the learned parameters}
In order to provide an extra analysis of the proposed method, we report the learned parameter values of the proposed similarity measure functions.
The values of the pretrained model are summarized in Table \ref{tab:learned_parameter}.
As we incorporate larger $n_k$, the learned parameters tend to be evenly distributed, which facilitates the diverse similarity measure functions to model the complex feature distributions.

\begin{table}
\centering
\caption{Learned parameter values of the proposed similarity measure functions with respect to various values of $n_k$.}
\resizebox{0.57\textwidth}{!}{\begin{tabular}{l|l|l} \toprule
 & $\lambda$ of the coarse network & $\lambda$ of the fine network \\ \toprule
$n_k=1$ & $1.15$ & $0.36$ \\
$n_k=2$ & $0.40, 1.59$ & $0.10, 0.41$ \\
$n_k=3$ & $0.18, 0.66, 2.19$ & $0.08, 0.39, 1.05$  \\
$n_k=4$ & $0.11, 0.41, 0.92, 2.68$ & $0.13, 0.51, 1.27, 3.67$ \\
$n_k=5$ & $0.08, 0.32, 0.52, 0.90, 2.36$ & $0.05, 0.32, 0.67, 1.48, 3.99$ \\
\bottomrule
\end{tabular}
}
\label{tab:learned_parameter}
\end{table}

\begin{figure*}[t]
    \centering
    \includegraphics[height=15cm]{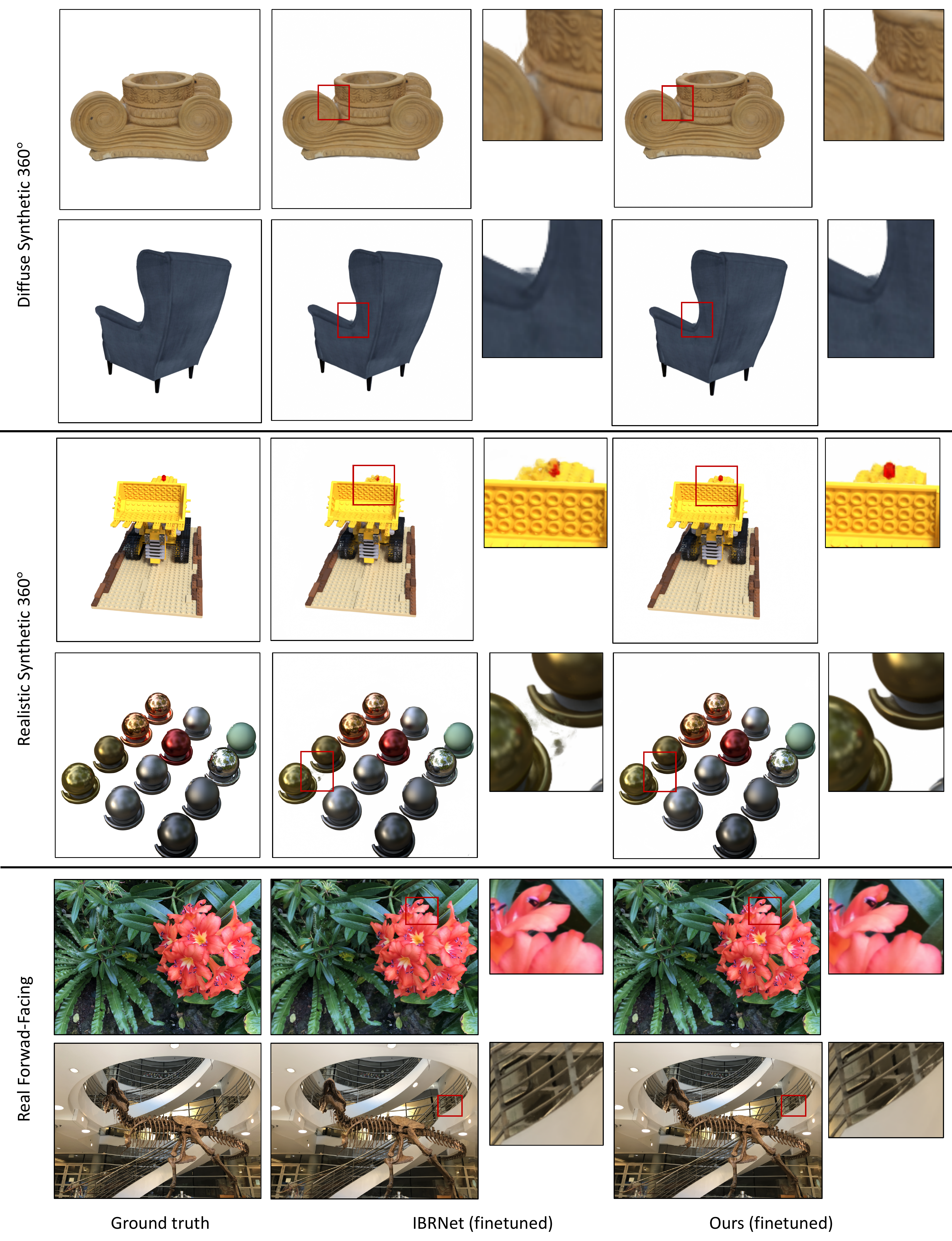}
    \caption{Qualitative comparison results on the benchmark datasets. The proposed method shows more accurate or comparable results compared with IBRNet.}
    \label{fig:qualitative_finetune}
\end{figure*}

\subsection{Additional qualitative results}

We provide qualitative comparison results with IBRNet \cite{wang:ibrnet:cvpr2021} based on the ``finetuned" protocol.
The comparison results on various scenes are visualized in Figure \ref{fig:qualitative_finetune}.
In the case of the Diffuse Synthetic 360$^{\circ}$ and Realistic Synthetic 360$^{\circ}$ datasets, the proposed method synthesizes the novel view images more accurately compared with IBRNet.
Specifically, in the first, second, and fourth cases, the proposed method shows better quality in edge regions.
In the third case, the proposed method synthesizes the ceiling region more clearly.
In the case of the Real Forward-Facing dataset, the proposed method shows comparable performances compared with IBRNet.
Note that we also submit the video comparison result on the ``pretrained" protocol.

\subsection{Additional ablation study}
\noindent
\textbf{Various aggregation methods.} For a further validation, we conduct ablation studies with different aggregation methods. For the experiments, we change the source feature aggregation method of IBRNet (element-wise mean+var aggregation) to element-wise mean aggregation (Exp1) and attention-based aggregation (Exp2). For the attention aggregation, we utilize Slot Attention \cite{locatello2020object} of GRF \cite{trevithick2021grf}. Experimental results (Exp1, Exp2, and Ours) in Table \ref{tab:ablation_2} show that our method is effective compared to different aggregation methods.\\\\
\textbf{Neighboring view selection.} The proposed scheme improves the baseline performance by handling occlusions from similar viewpoints, which is qualitatively shown in Figure 5 of the main paper. For a further validation, we conduct a comparison experiment of the baseline aggregation method and the proposed method in more harsher setting which does not select the most neighboring three views. In this setting, the occlusion might occur more frequently. The results are summarized in Table \ref{tab:ablation_2} (Exp3 and Exp4).
The proposed method shows the better performance, validating the occlusion handling efficacy of our method.

\begin{table*}[t]
\centering
\caption{Ablation results with respect to various settings on the benchmark datasets. For the protocol, ``P'' means the ``pretrained'' protocol. The best results are in \tb{bold}.}
\vspace{-2mm}
\resizebox{0.99\textwidth}{!}{%
\begin{tabular}{l|c|ccc|ccc|ccc} \toprule
\multirow{2}{*}{Method} & \multirow{2}{*}{Protocol} & \multicolumn{3}{|c}{Diffuse Synthetic 360$^{\circ}$ \cite{sitzmann:deepvoxels:cvpr2019}} & \multicolumn{3}{|c}{Realistic Synthetic 360$^{\circ}$ \cite{mildenhall:nerf:eccv2020}} & \multicolumn{3}{|c}{Real Forward-Facing \cite{mildenhall:llff:acm2019}} \\
\cline{3-11} & & PSNR$\uparrow$ & SSIM$\uparrow$ & LPIPS$\downarrow$ & PSNR$\uparrow$ & SSIM$\uparrow$ & LPIPS$\downarrow$ & PSNR$\uparrow$ & SSIM$\uparrow$ & LPIPS$\downarrow$  \\
\toprule
Exp1 (mean aggre) & P & 36.94 & 0.988 & 0.021 & 26.66 & 0.924 & 0.091 & 25.10 & 0.816 & 0.207 \\
Exp2 (atten aggre) & P & 36.92 & 0.987 & 0.022 & 26.82 & 0.920 & 0.105 & 25.12 & 0.813 & 0.207 \\
Exp3 (harsh; mean+var aggre) & P & 33.99 & 0.979 & 0.034 & 22.82 & 0.877 & 0.145 & 23.20 & 0.765 & 0.264 \\
Exp4 (harsh; our aggre) & P & 34.72 & 0.982 & 0.028 & 23.61 & 0.890 & 0.130 & 23.50 & 0.774 & 0.256 \\
\hline
Ours & P &  \tb{37.45} & \tb{0.990} & \tb{0.017} & \tb{27.13} & \tb{0.930} & \tb{0.083} & \tb{25.47} & \tb{0.826} & \tb{0.196} \\
\bottomrule
\end{tabular}%
}
\label{tab:ablation_2}
\end{table*}

\subsection{Additional implementation details}
We build the proposed framework based on the official IBRNet code\footnote{https://github.com/googleinterns/IBRNet}.
The proposed framework is trained on four V100 GPUs.
Each GPU utilizes a batch of $500$ rays which are sampled from a randomly selected scene.
For the feature extraction network, the coarse and fine networks share the same network except for the last layer, i.e., the coarse and fine networks have their own convolutional weights for the last layer.
In the feature extraction process, we only utilize the features extracted from the valid image region.
We also submit the code snippet of the proposed method for reference.

\end{document}